# Print2Volume: Generating Synthetic OCT-based 3D Fingerprint Volume from 2D Fingerprint Image

Qingran Miao, Haixia Wang, Haohao Sun, Yilong Zhang

*Abstract*—**Optical Coherence Tomography (OCT) enables the acquisition of high-resolution, three-dimensional fingerprint data, capturing rich subsurface structures for robust biometric recognition. However, the high cost and time-consuming nature of OCT data acquisition have led to a scarcity of large-scale public datasets, significantly hindering the development of advanced algorithms, particularly data-hungry deep learning models. To address this critical bottleneck, this paper introduces Print2Volume, a novel framework for generating realistic, synthetic OCT-based 3D fingerprints from 2D fingerprint image. Our framework operates in three sequential stages: (1) a 2D style transfer module that converts a binary fingerprint into a grayscale images mimicking the style of a Z-direction mean-projected OCT scan; (2) a 3D Structure Expansion Network that extrapolates the 2D image into a plausible 3D anatomical volume; and (3) an OCT Realism Refiner, based on a 3D GAN, that renders the structural volume with authentic textures, speckle noise, and other imaging characteristics. Using Print2Volume, we generated a large-scale synthetic dataset of 420,000 samples. Quantitative experiments demonstrate the high quality of our synthetic data and its significant impact on recognition performance. By pre-training a recognition model on our synthetic data and fine-tuning it on a small real-world dataset, we achieve a remarkable reduction in the Equal Error Rate (EER) from 15.62% to 2.50% on the ZJUT-EIFD benchmark, proving the effectiveness of our approach in overcoming data scarcity.**

*Index Terms*—**Optical Coherence Tomography, Volume Generation,**

## I. INTRODUCTION

Optical Coherence Tomography (OCT) [1] is a non-invasive, high-resolution 3D imaging technique that utilizes low-coherence light to capture 3D information within biological tissues at micrometer resolution. In recent years, OCT has been innovatively applied to the field of fingertip biometrics [2][3][4]. Unlike 3D fingerprints represented as point-cloud or meshes collected by contactless devices, which only contain epidermal information, OCT-based

This work was supported in part by Natural Science Foundation of Zhejiang Province under Grant LR24F030003, and National Natural Science Foundation of China under Grant 62376250, Grant 62276236, and Leading Innovation Team of Zhejiang Province under Grant 2021R01002. (Corresponding author: Haixia Wang.)

Qingran Miao is with the College of Information Engineering, Zhejiang University of technology, Hangzhou, China (e-mail: 2112112174@zjut.edu.cn)

Haixia Wang, Haohao Sun and Yilong Zhang are with the College of Computer Science and Technology, Zhejiang University of technology, Hangzhou, China (e-mail: hxwang@zjut.edu.cn; hhsun@zjut.edu.cn; zhangyilong@zjut.edu.cn).

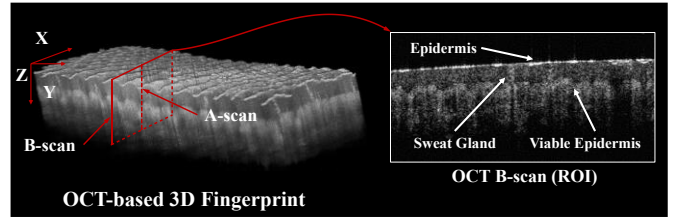

**Fig. 1.** OCT-based 3D fingerprint and fingertip skin structure collected in B-scan.

3D fingerprints consist of volume data that includes information from 0-3 millimeters beneath the fingertip skin [4][5] as shown in Fig. 1. From the epidermis layer within this volume data, the widely used External Fingerprint (EF) can be extracted [6][7]. The viable epidermis at the junction of the dermis and epidermis can be used to reconstruct the Internal Fingerprint (IF) [8][9]. The positions of sweat glands within the stratum corneum can be used to reconstruct sweat pore maps [10], which are difficult to capture from the skin surface [11][12]. Moreover, the morphology and distribution of the sweat glands themselves are considered to have significant identification potential [13]. Compared to traditional 2D fingerprints, OCT-based 3D fingerprints demonstrate significant advantages. Traditional 2D fingerprints are susceptible to interference from skin surface conditions (such as dirt, dryness, sweat, wear, or scars) and are easily deceived by spoof fingerprints. The internal fingerprint collected by OCT is hidden beneath the skin, making it effective against spoofing attacks and robust to various technical fingertip skin conditions.

However, the widespread of OCT application in biometrics still faces significant challenges. Current OCT devices are often bulky, expensive, and the acquisition process for 3D volume data is relatively time-consuming. These factors make large-scale collection of high-quality fingertip volume data extremely difficult and costly, thereby hindering the establishment of large public benchmark databases like FVC and NIST used in traditional 2D fingerprint research. The scale of existing public OCT fingerprint databases is generally limited, and this scarcity of data has become a key bottleneck restricting the development of the field. As summarized in table I, even the largest current database, ZJUT-EIFD [21], contains only 2,714 samples from 399 distinct fingers. This lack of data not only limits the development and validation of complex algorithms, especially data-hungry deep learning models, but also makes it difficult to fully evaluate the robustness and generalization capabilities. Consequently, OCT-based biometrics still has a long way to go before practical application.

TABLE I
CURRENT OCT-BASED 3D FINGERPRINT DATASETS

| Datasets | Year | Fingers | Samples | Technology |
|---|---|---|---|---|
| Bossen *et al.*[14] | 2010 | 51 | 153 | SS-OCT |
| Darlow *et al.*[15] | 2015 | 10 | 55 | SS-OCT |
| Aum *et al.*[16] | 2016 | 10 | 50 | SD-OCT[1] |
| SZU-OCTFD[17] | 2020 | 904 | 1808 | SD-OCT[2] |
| ZJUT-EFID[21] | 2023 | 399 | 2714 | SD-OCT[3] |

[1] Custom 1310nm spectral-domain OCT
[2] Custom 830nm SD-OCT at 18 KHz A-line rate
[3] Custom 1310nm SD-OCT at 60 KHz A-line rate

In contrast to the data scarcity in the 3D domain, synthetic 2D fingerprint generation is a relatively mature field, also developed to address the lack of large-scale, publicly available fingerprint datasets. Existing 2D synthesis techniques can be broadly divided into hand-crafted and learning-based approaches. Hand-crafted methods [22][23][24] often model fingerprint components such as orientation fields and minutiae separately, which can result in unrealistic ridge patterns and uniform ridge widths. More recent learning-based approaches [25][26][27][28][29] particularly those utilizing Generative Adversarial Networks (GANs), have significantly improved the realism of synthetic fingerprints.

Despite these advancements in 2D synthesis, there are currently no established strategies for generating synthetic OCT-based 3D fingerprint. The transition from synthesizing a 2D surface image to a full 3D biometric volume presents unique and significant challenges. First, it requires extending a 2D fingerprint pattern into a complex three-dimensional structure with depth information, algorithmically creating plausible and consistent subsurface layers like the stratum corneum and viable epidermis junction. Second, rendering this volume data to be realistic is exceptionally difficult; it involves simulating the specific textural properties, speckle noise, and signal attenuation characteristic of the OCT imaging process to ensure the synthetic data has a similar domain to real-world scans.

Therefore, this paper proposes Print2Volume, a novel framework capable of generating a virtual OCT-based 3D Fingerprint by recovering deep fingertip skin information from a single 2D fingerprint image. Print2Volume comprises three stages. First, a virtual binarized fingerprint (generated by PrintsGAN [29]) is style-transferred to resemble a Z-direction mean fingerprint from an OCT volume, providing the necessary fingerprint template and grayscale information for subsequent 3D generation. Then, a 3D structure expansion network is utilized to recover the deep structure from the 2D fingerprint. Finally, we employ a GAN with a 3D discriminator to render the volume data with realism, producing the final virtual OCT-based 3D Fingerprint. In our experiments, we utilize Print2Volume to create a large-scale synthetic dataset containing 420,000 OCT-based 3D Fingerprint samples from 28,000 distinct identities with 15 impressions each. We demonstrate how this synthetic data can be used to train a deep network to extract a fixed-length representation for recognition. Specifically, by pre-training a network on the database generated by Print2Volume and then fine-tuning it on a small, real-world OCT dataset (320 fingers, 8 impressions each), we achieve a significant improvement in True Acceptance Rate (TAR) and Equal Error Rate (EER) on the real fingerprint evaluation dataset, ZJUT-EIFD, compared to training only on the limited real data.

Our contributions can be summarized as follows:

- We propose Print2Volume, the first-ever framework designed to generate complete and realistic OCT-based 3D fingerprint volumes from a single 2D fingerprint image.
- We design a novel and effective three-stage pipeline that systematically recovers the 3D structure and renders it with high perceptual realism, successfully bridging the domain gap from 2D prints to 3D OCT volumes.
- We demonstrate the capability of our framework by creating a large-scale synthetic OCT fingerprint dataset, which is crucial for advancing research in this data-scarce field.
- We validate through extensive experiments that pre-training on our synthetic data significantly boosts the performance of recognition models, drastically reducing error rates compared to training on limited real data alone.

## II. Synthetic OCT-based 3D Fingerprint Generation

This section first introduces the overview of proposed Print2Volume, followed by 2D OCT Z-direction mean fingerprint generation, the details of proposed 3D structure expansion network and OCT refiner GAN.

### A. Overview

Our proposed framework, Print2Volume, is designed to recover a complete and realistic OCT-based 3D fingerprint volume from a single 2D fingerprint image. The entire process is structured into three distinct yet sequential stages: (1) 2D Style Transfer, (2) 3D Structure Expansion, and (3) OCT Realism Refiner. Fig. 2 illustrates the overall pipeline.

First, Print2Volume begin with a synthetic 2D binary fingerprint $I_{ID} \in \{0,1\}^{256 \times 256}$ which is generated via a random noise vector $z_{ID} \in \mathbb{R}^{512}$, which defines the unique ridge and valley pattern as a new fingerprint identity. Next, $I_{ID}$ along with a warping noise vector $z_{distort} \in \mathbb{R}^{16}$ is passed to a non-linear Thin-Plate-Splinem (TPS) warping module and cropping GAN to produce a warped Master-Print $I_M$ . However, $I_M$ lacks the specific grayscale characteristics of OCT scans. To bridge this domain gap, the 2D Style Transfer stage uses a generator $G_S(\cdot)$ to transform $I_M$ into a grayscale image $I_S$ that mimics the style of a Z-direction mean-projected OCT image. This intermediate image $I_S$ serves as a crucial template, providing both the foundational fingerprint pattern and plausible grayscale information for the subsequent depth recovery.

Second, the 3D Structure Expansion stage takes the style-transferred 2D fingerprint $I_S$ as input. A dedicated 3D expansion network $G_E(\cdot)$ is employed to extrapolate this 2D infor-mation into the depth dimension (Z-axis). This network is trained to predict a structural volume $V_E$ by generating consistent and plausible subsurface layers, effectively transforming the flat fingerprint pattern into a three-dimensional anatomical structure.

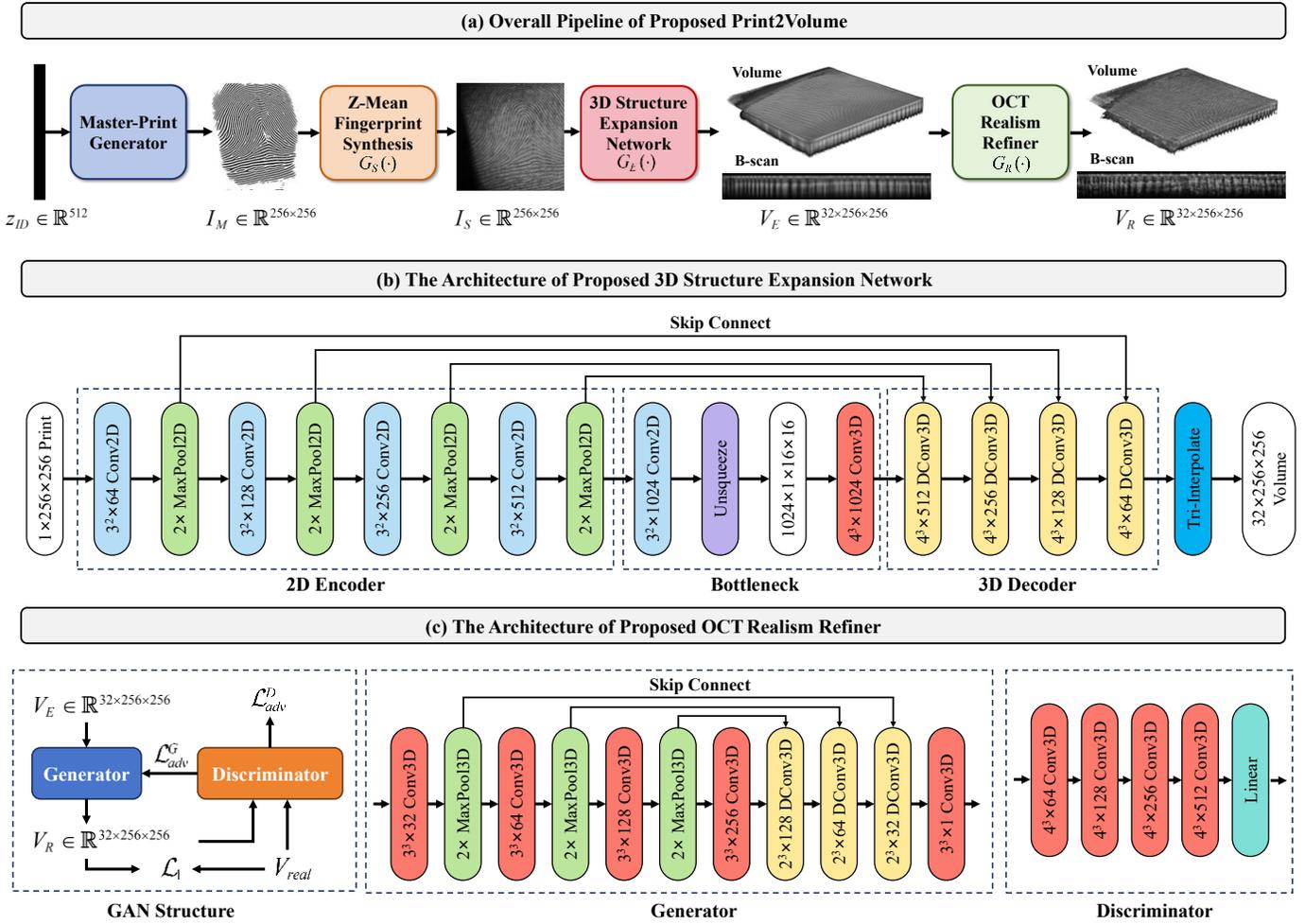

**Fig. 2.** Detail of proposed Print2Volume. (a) Overall pipeline of proposed Print2Volume, (b) The Architecture of Proposed 3D Structure Expansion Network, and (c) The Architecture of Proposed OCT Realism Refiner.

Finally, $V_E$ lacks the realistic textural details and noise profiles inherent in real OCT scans. We proposed OCT Realism Refiner $G_R(\cdot)$ takes $V_R$ and refines it by adding authentic OCT-like textures, speckle noise, and signal attenuation effects. A 3D discriminator ensures the perceptual quality and authenticity of the final output by learning to distinguish between the generated volumes and real OCT data. The result of this stage is the final, realistic virtual OCT-based 3D fingerprint.

### B. Z-Mean Fingerprint Synthesis

To generate a foundational fingerprint identity, we learn a mapping from a random noise vector $z_{ID} \in \mathbb{R}^{512}$ to a binary print $I_{ID} \in \{0,1\}^{256 \times 256}$ inspired by PrintsGAN. We utilize a GAN for this task, where the generator $G_I(\cdot)$ creates the synthetic binary fingerprint and the discriminator $Disc_{ID}$ aims to distinguish it from binarized real fingerprints. The training is guided by the classic adversarial loss function

$$
\begin{aligned}
\mathcal{L}_{adv}(G_I, Disc_{ID}) = {} & \mathbb{E}_I[\log Disc_{ID}(I)] \\
& + \mathbb{E}_z[\log(1 - Disc_{ID}(G_I(z)))].
\end{aligned}
\tag{1}
$$

Once a unique Master-Print $I_{ID}$ is generated, it undergoes a non-linear distortion and cropping process to simulate the various impressions of a single finger being placed on a scanner. This is achieved using a dedicated GAN that applies a Thin-Plate-Spline (TPS) warping transformation to the Master-Print, resulting in a realistically warped and cropped binary fingerprint $I_M$. By fixing the identity vector $z_{ID}$ while varying the distortion parameters, we can generate multiple, unique impressions for a single fingerprint identity.

$I_M$ provides the necessary structural pattern but lacks the grayscale texture essential for the subsequent 3D structure expansion. To address this, we perform an exemplar-based style transfer to convert $I_M$ into the style of a Z-direction mean-projected OCT fingerprint. We adopt MATEBIT [30] as $G_S(\cdot)$ to endow the $I_M$. $I_M$ as the content input and a real Z-mean OCT fingerprint as the style exemplar $T_{style}$, generating a high-fidelity grayscale template $I_S$ . To ensure the generated image is stylistically consistent with the exemplar on a global level, the framework is guided by a Contrastive Style Learning (CSL), which is defined as:

$$\mathcal{L}_{style} = -\log \frac{\exp(\frac{z^T z^+}{\mathcal{T}})}{\exp(\frac{z^T z^+}{\mathcal{T}}) + \sum_{j=1}^{m} \exp(\frac{z^T z_j^-}{\mathcal{T}})}, \qquad (2)$$

where $z$ is the style code of the generated image (anchor), $z^+$ is the code of the positive sample (exemplar image), and $z_j^-$ are the codes of the negative samples. The learned high-quality global style code is then injected into the network's decoder via Adaptive Instance Normalization (AdaIN) layers to control the overall style. The entire network is optimized end-to-end using multiple objective functions, including adversarial, contextual, and CSL. This joint optimization ensures that the resulting $I_S$ faithfully reproduces the style of an OCT image while preserving the original structure of $I_M$ laying a robust foundation for the subsequent 3D reconstruction. Notably, the style transfer stage requires $T_{style}$ to serve as style exemplars. Therefore, to ensure rendering quality and textural richness, we classify the input fingerprints into four categories based on their foreground position: upper, middle, lower, and full. A dedicated pool of 100 real exemplars is provided for each category.

## C. 3D Structure Expansion Network

The core task of the second stage is to recover a 3D structural volume $V_E \in \mathbb{R}^{1 \times D \times H \times W}$ from $I_S$ . To accomplish this, we propose a 3D structure expansion network $G_E(\cdot)$ , which takes a single 2D image as input and outputs a corresponding 3D volume. The network architecture consists of three main components as shown in Fig. 2(b): a 2D Encoder, a Bottleneck for dimensional transition, and a 3D Decoder. The encoder extracts ridge flow and pore semantics in X-Y, while the decoder progressively expands and refines along Z-axis. This preserves lateral fidelity, yields physically plausible stratified structure.

Concretely, we use a four-level U-Net. The encoder stacks ConvBlock2D (Conv–InstanceNorm2d–ReLU)×2 with 2× max-pooling, expanding channels up to 16×64 and downsampling to $H/16 \times W/16$ . At the bottleneck, we first aggregate semantics with 2D Convs, then unsqueeze a unit depth dimension and apply a 3×3×3 Conv with InstanceNorm3d to accomplish a seeded 2D-to-3D lift. The decoder performs four stages of ConvTranspose3d for learnable upsampling in $(D, H, W)$ . Each stage concatenates a skip connection lifted from the encoder. To align 2D skips to 3D, we repeat features along depth and use trilinear interpolation to match the current decoder scale, injecting stable lateral cues into the growing volume. Because transposed convolutions can introduce one-voxel size drift, we add a final trilinear interpolation to the exact target $(D, H', W')$ , followed by a 1×1×1 Conv and Sigmoid to produce a [0,1]-normalized structural volume (typical $D = 32$ , $H' = W' = 256$ ). Training uses voxel-wise BCE add 3D-SSIM to further reinforce intensity and structural consistency. The final objective is

$$\mathcal{L}_E = BCE(V_{real}, G(I_s)) + \text{3D-}SSIM(V_{real}, G(I_s)). \qquad (3)$$

In practice, this design concentrates capacity where the information is richest (lateral encoding), then injects the necessary depth interactions and layering during decoding, producing 3D structures that are later refined in the realism stage.

## D. OCT Realism Refiner

The structural volume $V_E$ produced by the expansion network correctly outlines the anatomical shape of the fingerprint layers, but it lacks the realistic textural details, speckle noise, and other artifacts inherent in real-world OCT scans. To address this, the final stage of our framework employs an OCT Realism Refiner, which is a 3D Generative Adversarial Network (GAN) designed to translate the clean structural volume into a perceptually authentic OCT volume $V_R$ . This GAN consists of a 3D U-Net Generator and a 3D PatchGAN Discriminator.

The generator $G_R(\cdot)$ is responsible for rendering the input structural volume with realistic details. Its architecture is a full 3D U-Net. The encoder path consists of three ConvBlock3D (3DConv–BatchNorm3d–ReLU), each followed by a max-pooling layer, to progressively downsample the volume and extract hierarchical features. The bottleneck further processes this compact representation. The decoder path then uses 3D transposed convolutional layers (ConvTranspose3D) to upsample the features back to the original volume size $V_R \in \mathbb{R}^{1 \times D \times H \times W}$ . To ensure the generated volumes are realistic, we employ a 3D PatchGAN discriminator $D_R(\cdot)$ . Instead of classifying the entire volume as real or fake with a single score, this discriminator provides feedback on a patch-by-patch basis, which encourages the generator to produce high-frequency details more effectively. The discriminator takes a 3D volume as input and processes it through a series of four strided 3D convolutional layers, each followed by a LeakyReLU activation. This architecture progressively reduces the spatial dimensions of the input, ultimately producing a 3D tensor where each element in the tensor corresponds to the realness of a specific patch in the original input volume. Training follows a standard conditional adversarial objective with a reconstruction prior. Loss can be expressed as

$$\mathcal{L}_{adv}^{G} = \mathbb{E}\left[\log(1 - D_R(G_R(V_E)))\right], \qquad (4)$$

and

$$\mathcal{L}_{adv}^{D} = \mathbb{E}\left[\log(D_R(V_{real}))\right] + \mathbb{E}\left[\log(1 - D_R(G_R(V_E)))\right]. \qquad (5)$$

The refiner additionally minimizes an L1 loss fidelity term $\mathcal{L}_1 = \left\| \tilde{V}_{real}, V_R \right\|$ to anchor global intensity and layer boundaries. The final objective is

$$\mathcal{L}_R = \mathcal{L}_{adv}^{G} + \alpha \mathcal{L}_1, \mathcal{L}_D = \mathcal{L}_{adv}^{D}. \qquad (6)$$

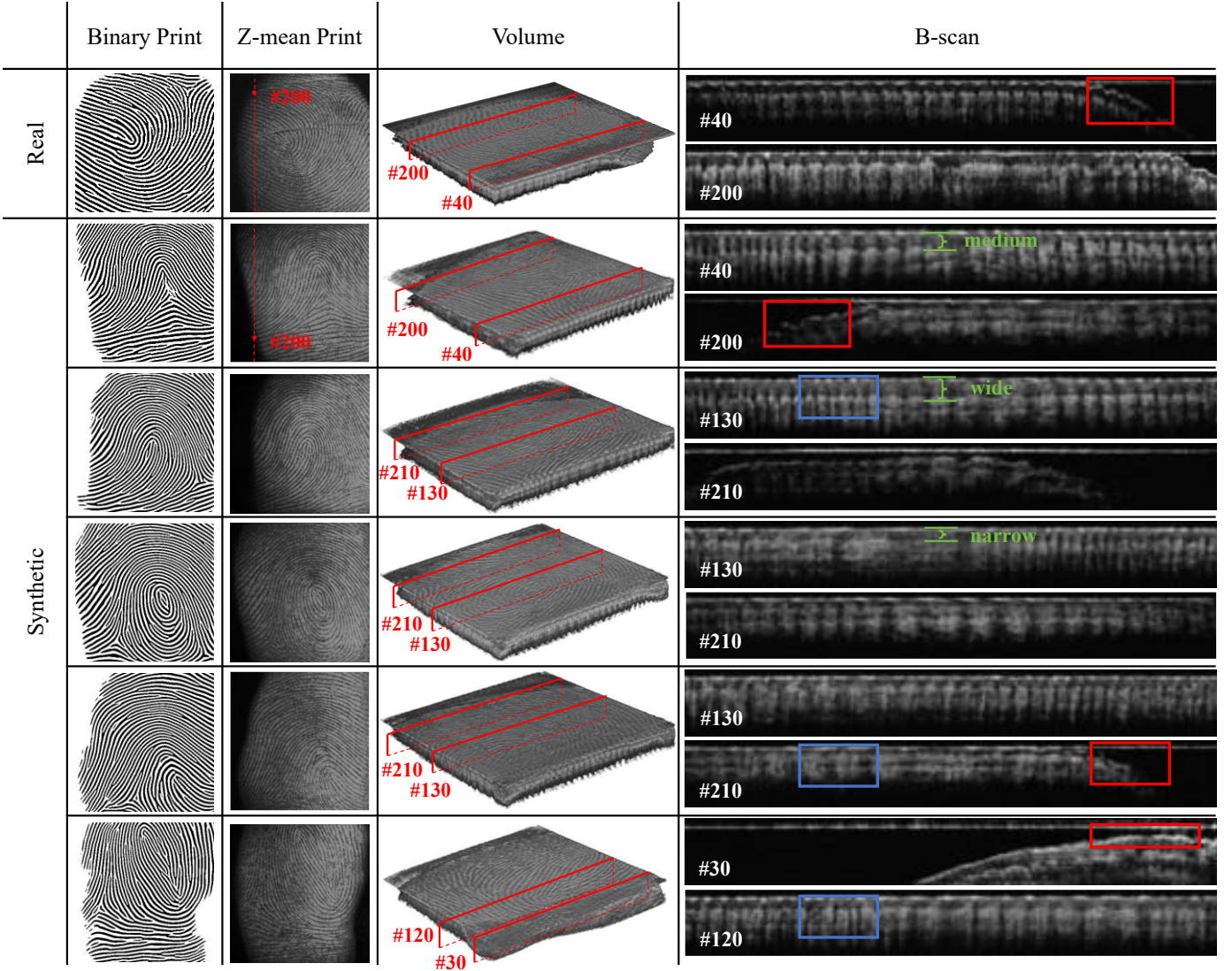

**Fig. 3.** Comparison of Synthetic OCT-based 3D fingerprint generated by Print2Volume and real OCT volume.

## III. Experiment

In this section, we first present the detailed experimental setup and the construction of the synthetic OCT-based 3D fingerprint dataset. We then evaluate the generated data both qualitatively and quantitatively.

### A. Implementation Details

The methodology was implemented using the PyTorch framework, and all models were trained on four NVIDIA A6000 GPUs. The training data for our method were drawn from 2,714 fingertip OCT samples provided by the ZJUT-EIFD dataset. The training process for each part is detailed as follows. For the Z-Mean fingerprint synthesis, MATEBIT is trained on a paired dataset of $z$-axis mean (en-face) fingerprints and binarized fingerprints from ZJUT-EIFD. The binarized fingerprints are extracted using the VeriFinger SDK [32]. The rest of the settings follow the original paper [30]. The 3D Structure Expansion Network is trained on paired Z-axis mean fingerprints and real volume. For training the OCT

Realism Refiner, we first use the pre-trained $G_E(\cdot)$ to generate VE from the z-axis mean fingerprint, then train the refiner on paired $V_E$ and real volume. $\alpha$ in eq.(6) is set to 10. The two networks proposed in this paper use the Adam optimizer with a learning rate of 2e-4 and train for 40 and 60 epochs respectively.

For the creation of the synthetic OCT-based 3D fingerprint dataset, we reduced costs by directly using the 2D fingerprints generated by PrintsGAN [29] as masterprint templates. Specifically, we first trained an auto-encoder by L2 loss to learn the mapping from a raw fingerprint to a binary fingerprint using 10K groundtruth binary fingerprints. Then we used it to rapidly convert the PrintsGAN fingerprints into binarized fingerprints, resulting in 28,000 classes and a total of 420,000 masterprints (15 per class). Next, we applied the trained modules $G_S(\cdot)$, $G_E(\cdot)$, and $G_R(\cdot)$ to transform all these fingerprints into OCT-based 3D fingerprints, producing the final datasets.

| | Z-mean Print #Template | Triplanar Slice View | En-Face Fingerprint | | Z-mean Print #Generate | SSIM |
|---|---|---|---|---|---|---|
| | | | External | Internal | | |
| Synthetic | 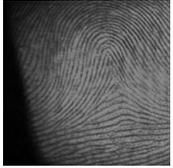 | 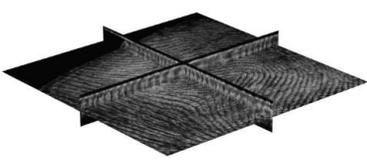 | 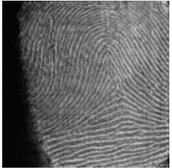 | 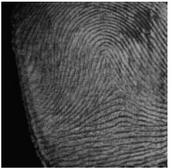 | 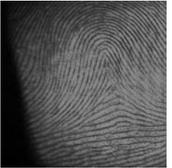 | 0.9245 |
| | 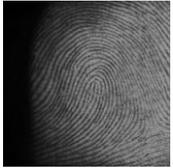 | 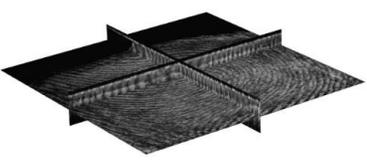 | 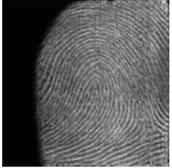 | 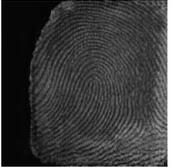 | 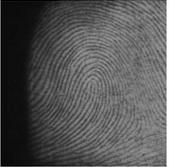 | 0.8914 |
| | 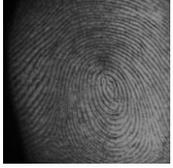 | 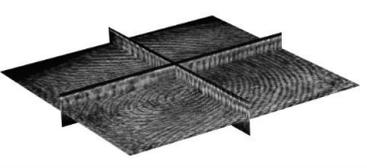 | 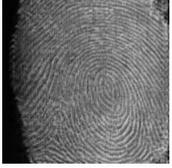 | 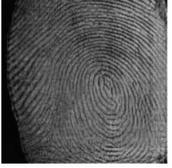 | 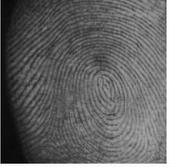 | 0.9124 |
| | 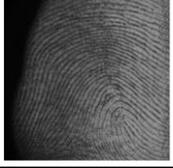 | 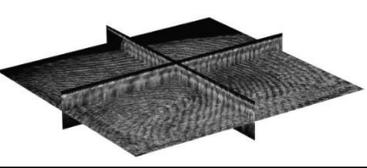 | 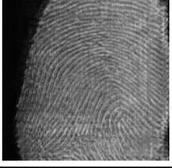 | 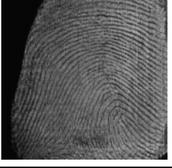 | 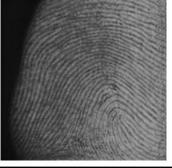 | 0.8957 |
| | 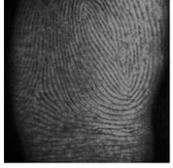 | 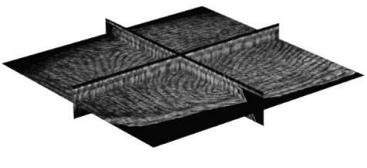 | 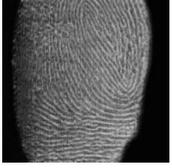 | 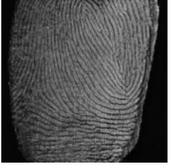 | 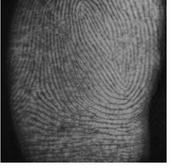 | 0.9046 |

**Fig. 4.** Evidence for generated volume data faithfully restored the fingerprint texture.

### B. Qualitative Results

Qualitatively, the synthetic OCT-based 3D fingerprints generated by Print2Volume are both highly realistic and discriminative.

Fig. 3 shows a detailed comparison between real OCT volume data and the synthetic data generated by Print2Volume. As can be observed, the Z-mean print produced by our method is not only stylistically similar to the real data but also accurately restores the ridge and valley fluctuations of the fingerprint, providing a solid foundation for the subsequent generation of 3D volume data. A comparison of the B-scan images from real and virtual data reveals that our method can successfully recover the gaps that appear in real OCT data when the fingertip epidermis is not in full contact with the glass plate, based on the grayscale information of the 2D fingerprint (as shown in the red boxes). Notably, the #30 B-scan in the sixth row demonstrates that the method can even estimate the state of the B-scan when the galvanometer has not yet moved into position and the fingertip is completely separated from the glass plate, a situation that often occurs at the beginning of real data acquisition. Furthermore, the green brackets and blue boxes indicate that Print2Volume is capable of generating stratum corneum of varying widths and diverse types of dermal layers, which significantly enhances the diversity of the synthetic data.

To verify whether the generated volume data faithfully restored the fingerprint texture, we conducted a multi-faceted analysis as shown in Fig. 4. We extracted the external fingerprint from the epidermal layer, the internal fingerprint from the viable epidermis, and the Z-mean fingerprint. Visually, the texture information of all three is highly consistent with the 2D fingerprint template. We also calculated the Structural Similarity Index (SSIM) between the generated Z-mean fingerprint and the template. The results show a high degree of similarity, with minor differences primarily arising from the depth information introduced during the 3D expansion, which affects the mean grayscale values after projection. The fundamental structure remains consistent.

Furthermore, Fig. 5 presents a comparison between the structural volume generated by the Expansion Structure Network and the result after processing by the OCT Realism Refiner. An interesting observation is that the realism provided by the GAN is, in fact, a form of degradation from a structural clarity standpoint. It enhances perceptual realism by introducing artifacts inherent to OCT imaging, such as speckle noise and signal attenuation. From the perspective of pure fingerprint quality, both the internal and external fingerprints extracted directly from the output of the Expansion Structure

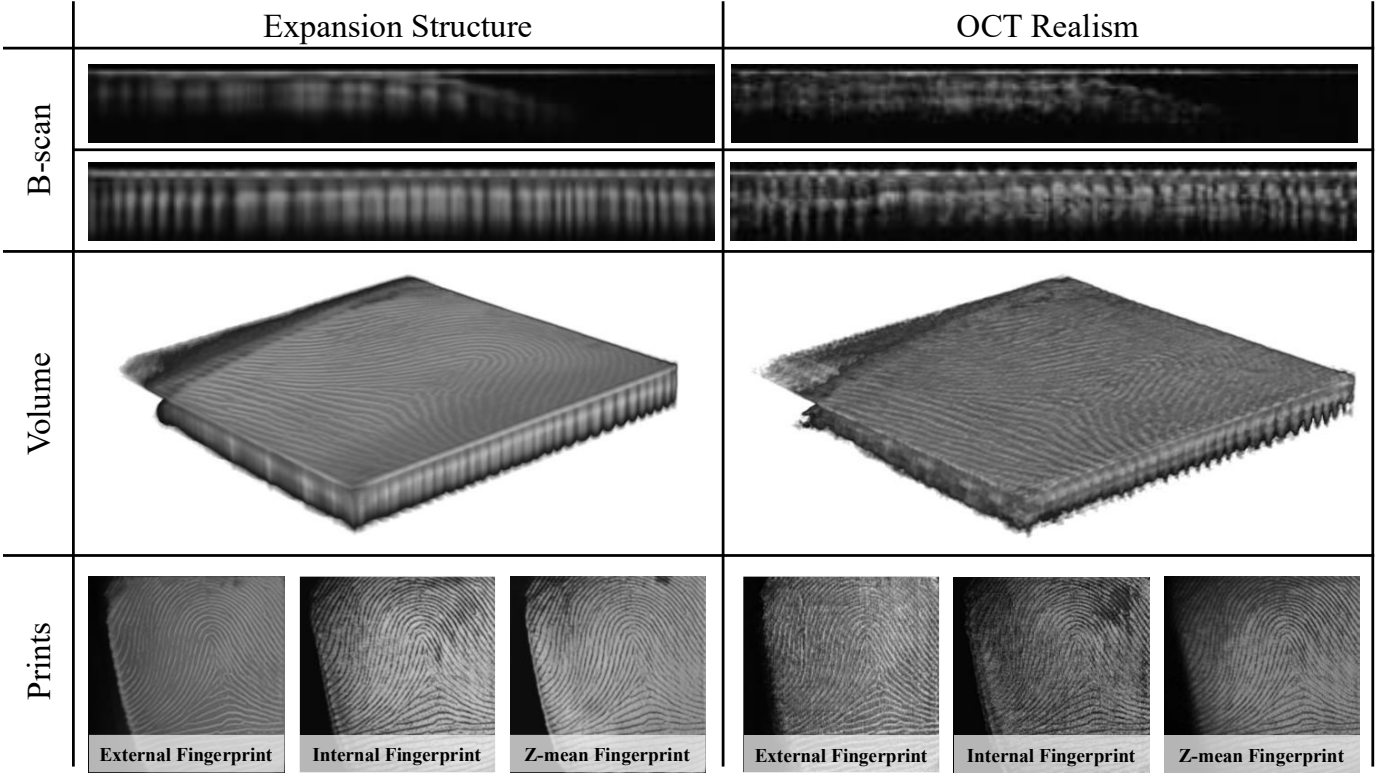

**Fig. 3.** Comparison of Synthetic OCT-based 3D fingerprint generated by Print2Volume and real OCT volume.

Network are of higher quality and clarity. This finding suggests a novel application for our method: it can not only serve as a data augmentation technique, but its intermediate structural volumes can also function as high-quality labels for research in subcutaneous biometric feature extraction. In practice, real-world OCT volumes are often corrupted by speckle noise and other artifacts, and it is not feasible to directly acquire high-quality en-face fingerprints. Obtaining clear internal fingerprints typically requires laborious manual annotation of the dermal layer contours. This difficulty in label creation has become a significant bottleneck hindering the development of the field. In contrast, the volumes generated by our method originate from clean, binarized fingerprints, and the B-scans produced by the Expansion Structure Network can serve as effective labels for the dermal and epidermal layer contours. This would greatly alleviate the demand for manual annotation, thereby advancing research in this area.

### C. Quantitative Evaluation

For the quantitative analysis, as there are no other OCT-based 3D fingerprint generation methods available for comparison to the best of our knowledge, we primarily evaluate the quality of the data generated by Print2Volume in two aspects: reporting the distance between the generated data and real data, and the performance improvement that the synthetic dataset brings to a recognition task.

First, we employ the Fréchet Video Distance (FVD) and Fréchet Inception Distance (FID) to evaluate the perceptual quality of the generated volumes and B-scans, respectively.

FVD quantifies the dissimilarity between generated and real volumes by extracting image features with the I3D model [33], which is well-suited for videos, denoted as $FVD_{I3D}$. FID assesses the quality of generated images, but at a slice-level using the InceptionV3 model [34]. The evaluation was conducted using 30,000 generated volumes and their corresponding B-scans. The results are presented in Table II.

The results clearly demonstrate the effectiveness of the OCT Realism Refiner in enhancing the authenticity of the generated data. The $FVD_{I3D}$ score, which measures the perceptual quality of the entire volume, decreased substantially from 2945.5 for the initial structural volume to 1564.6 for the refined volume. Similarly, the slice-level FID score improved, dropping from 94.15 to 67.45. Lower scores in both metrics indicate that the distribution of the generated data is closer to that of real data. This quantitatively confirms that the final refiner stage successfully imparts the necessary realistic textures, speckle noise, and other characteristics, making the synthetic volumes perceptually much closer to real OCT scans.

Next, we test the impact of the synthetic dataset on the recognition performance of a fixed-length representation model for OCT-based 3D fingerprint. In this experiment, the ZJUT-EIFD dataset serves as the test set. We use another self-build datasets containing 2,560 volumes (320 distinct fingers with 8 samples each) as the training or fine-tuning real set. We report the results for three different training strategies: training on the self-build real dataset alone, training on the synthetic dataset alone, and pre-training on the synthetic dataset followed by fine-tuning on the real dataset. The results are re-ported in Table III.

TABLE II
RESULTS ON PERCEPTUAL QUALITY EVALUATION

| Metric | $FVD_{I3D}$ | FID |
|---|---|---|
| 3D Structure Expansion | 2945.5 | 94.15 |
| OCT Realism Refine | **1564.6** | **67.45** |

TABLE III
RESULTS ON IMPROVED RECOGNITION PERFORMANCE AT
EQUAL ERROR RATE (%) AND TRUE ACCEPT RATE (%)

| Metric | EER | TAR[1] | TAR[2] |
|---|---|---|---|
| Self-build training | 15.62 | 78.95 | 51.35 |
| Synthetic training | 3.51 | 98.61 | 91.92 |
| Synthetic training + Self-build fine-tuning | **2.50** | **99.12** | **95.10** |

[1] TAR result @ FAR=0.1%.
[2] TAR result @ FAR=0.01%

Table III showcases the significant impact of our synthetic data on recognition performance. Training the model solely on the limited self-build real dataset results in a relatively high Equal Error Rate (EER) of 15.62% and True Acceptance Rates (TAR) of 78.95% (TAR[1] @ FAR=0.1%) and 51.35% (TAR[2] @ FAR=0.01%). This baseline performance highlights the challenge of training deep models with insufficient real data.

Remarkably, when the model is trained exclusively on our large-scale synthetic dataset, the performance improves dramatically. The EER plummets to just 3.51%, while the TARs surge to 98.61% and 91.92% respectively. This demonstrates that the data generated by Print2Volume is not only realistic but also contains rich, discriminative features that are highly effective for training a robust recognition model, even outperforming a model trained on a small set of real data.

The most significant performance gain is achieved by leveraging the synthetic data for pre-training and then fine-tuning the model on the real dataset. This strategy yields the best results across all metrics, achieving the lowest EER of 2.50% and pushing the TARs to their highest values of 99.12% and 95.10%. This outcome strongly supports our central hypothesis: large-scale, high-quality synthetic data generated by Print2Volume can effectively overcome the data scarcity problem in the OCT-based fingerprint domain, significantly boosting the accuracy and reliability of recognition systems.

## IV. CONCLUSION

In this paper, we addressed the critical issue of data scarcity in the field of OCT-based 3D fingerprint recognition. We introduced Print2Volume, a novel and effective framework capable of generating realistic synthetic OCT volumes from readily available 2D fingerprint images. Our three-stage approach—encompassing 2D style transfer, 3D structure expansion, and GAN-based realism refinement—successfully transforms a simple 2D pattern into a complex and authentic 3D biometric sample.

Our qualitative results demonstrated that the generated data is visually similar to real OCT scans, accurately capturing anatomical features such as the stratum corneum, dermal-epidermal junction, and even common imaging artifacts. The quantitative analysis further confirmed the high quality of our synthetic data. Perceptual metrics showed a significant improvement in realism, and more importantly, recognition experiments proved its practical utility. By leveraging our synthetic data for pre-training, we achieved a state-of-the-art recognition performance on a real-world benchmark, reducing the EER by over 80% compared to a model trained on limited real data. This outcome provides strong evidence that our method can effectively mitigate the data scarcity problem and unlock the potential of deep learning for this domain.

Furthermore, we highlighted the potential of the intermediate structural volumes generated by our network to be used as clean labels for subcutaneous biometric research, offering a path to bypass the laborious process of manual data annotation. Future work could explore expanding the diversity of the generated anatomical structures and adapting the Print2Volume framework to other volumetric medical imaging modalities that face similar data scarcity challenges.